\documentclass[letterpaper, 10 pt, conference]{ieeeconf}  
\usepackage{booktabs}
\usepackage{tabularx}
\usepackage{graphicx}
\usepackage{multirow}
\usepackage{makecell}
\usepackage{csquotes}

\IEEEoverridecommandlockouts                              

\usepackage{amsmath} 
\usepackage{amssymb}  
\usepackage{url}

\title{\LARGE \bf
Enforcing Human-like Kinematics in Dexterous Piano Playing via Adversarial Posture Regularization
}

\author{
\begin{tabular}{c}
Bin Qiu$^{1}$, Yanming Shao$^{2,3}$, Guanyu Cai$^{1}$, and Yao Mu$^{1,*}$\\[1.5mm]
{\footnotesize
$^{1}$\,Shanghai Jiao Tong University
\qquad
$^{2}$\,The University of Hong Kong
\qquad
$^{3}$\,Shanghai AI Laboratory
\qquad
$^{*}$\,Corresponding author
}
\end{tabular}
}

\begin{document}

\maketitle
\thispagestyle{empty}
\pagestyle{empty}

\begin{abstract}
Reinforcement learning can train bimanual dexterous hands to play piano in physics simulation with high note accuracy, but for high-DoF dexterous hands, relying solely on task rewards or IK inversion often leads to unnatural postures and joint overextension. We propose \textit{Adversarial Posture Regularization (APR)}. It avoids expensive, song-aligned expert demonstration data and instead uses a small amount of casual human playing data. By matching the distribution of the posture of the policy with the human prior through an adversarial objective, APR encourages more human-like hand shapes. Meanwhile, we collect and release unstructured hand motion data of piano playing using a consumer-grade Meta Quest 3, and retarget the key motion information to the Shadow Hand. Finally, we achieve significantly better performance than prior methods on all three human-likeness metrics (cPSI, BSE, and FAC) as well as in visual quality. Project repository: \url{https://github.com/APRProject/APRPianist}.

\end{abstract}

\section{INTRODUCTION}
Humans can dexterously use both hands to perform music with graceful, fluid motions. In contrast, although bimanual manipulation has become increasingly capable in recent years, with systems demonstrating coordinated behaviors in tasks such as heavy object co-transport~\cite{dio2023cooperative} and garment folding~\cite{avigal2022speedfolding}, scaling these successes to human level control with two dexterous hands remains unresolved. Piano playing offers a particularly stringent benchmark, combining continuous spatial accuracy with fast temporal dynamics and requiring tightly coupled spatiotemporal coordination over more than 40 degrees of freedom (DoFs). Recent works such as Google's RoboPianist~\cite{zakka2023robopianist} demonstrate that agents can learn to play the piano via deep reinforcement learning in physics simulators like MuJoCo~\cite{todorov2012mujoco} by leveraging task rewards.

Despite this impressive progress, most learning-based methods evaluate dexterous manipulation primarily through task-centric metrics, such as the F1-score, which only reflect keystroke accuracy and timing. Optimizing purely for these sparse task rewards invariably encourages policies to exploit simulator dynamics, leading to severe ``reward hacking"~\cite{skalse2022defining} such as extreme joint hyperextension to avoid accidental keystrokes.

Furthermore, although some recent methods attempt to alleviate this issue by incorporating classical inverse kinematics (IK)~\cite{caron2024pink} together with fingertip based 2D planar supervision~\cite{qian2025pianomime,zhang2020mediapipe}, the massive kinematic redundancy of high-dimensional hands, for example the 24-DoF Shadow Hand, means that constraining only endpoint positions leaves the proximal joints severely under constrained. As a result, to satisfy rigid fingertip targets, IK solvers and tracking driven policies often force intermediate joints, such as the metacarpophalangeal (MCP) and proximal interphalangeal (PIP) joints, into collapse like bending and contortions, producing biomechanically implausible “zombie hand” configurations. 

Intuitively, imitation learning could provide an effective way to constrain full hand~\cite{rajeswaran2017learning}. However, obtaining full hand pose trajectories for complete musical pieces via accurate spatial tracking is often prohibitively expensive. Such data typically requires skilled performers, carefully calibrated capture rigs, and controlled recording sessions, making large scale collection difficult to scale. Overall, the field still lacks a robust mechanism that can smoothly constrain the high-dimensional posture space during highly dynamic execution, without relying on large scale, high fidelity full pose tracking data.

\begin{figure}[t] 
    \centering 
    \includegraphics[width=\columnwidth]{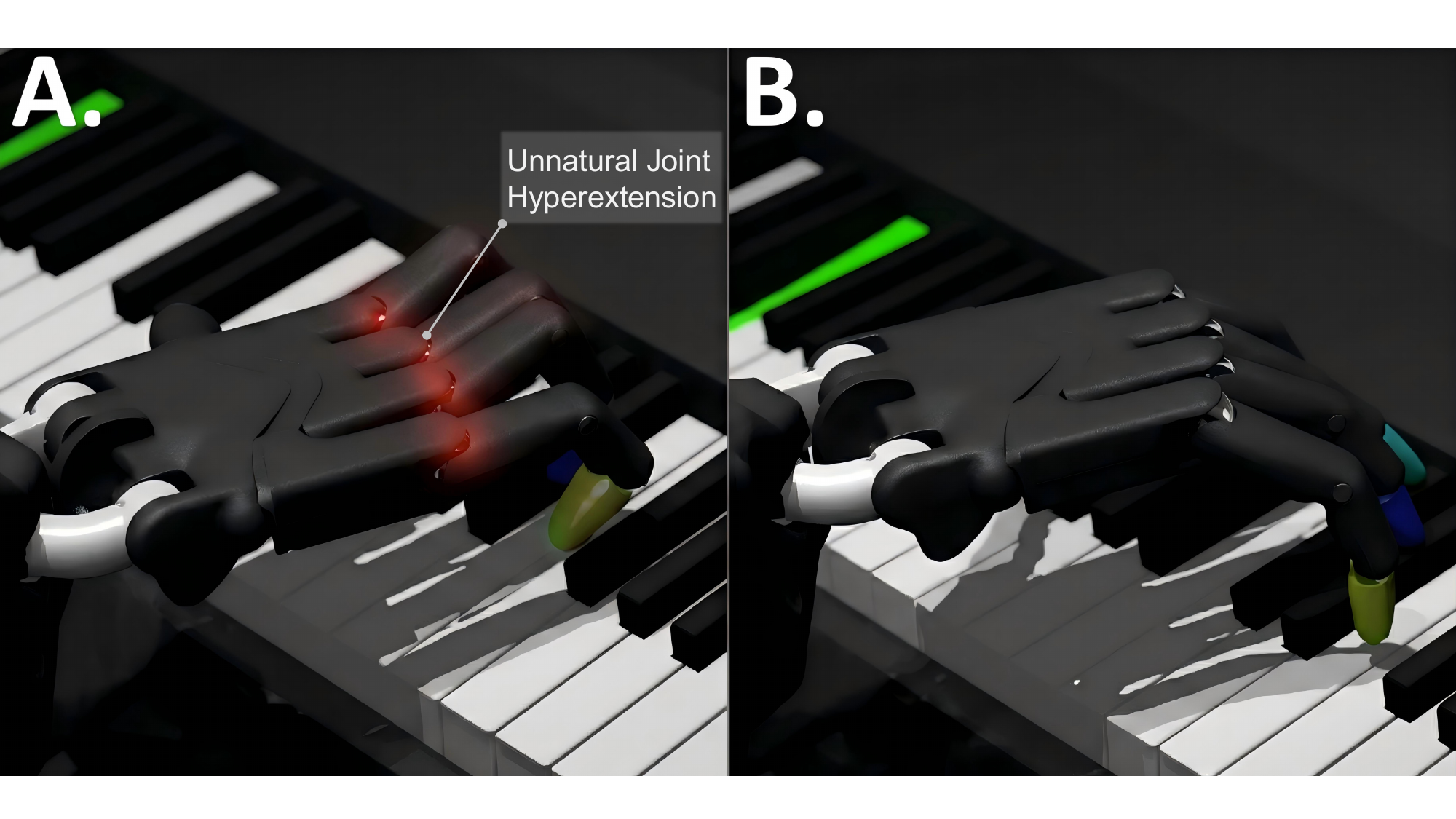} 
    \caption{Qualitative comparison of hand postures. \textbf{(A) Baselines:} Without full posture constraints, the policy shows severe biomechanical distortions and “zombie hand” artifacts. Red highlights mark unnatural hyperextension of non-playing fingers. \textbf{(B) Ours (APR):} APR softly constrains finger postures. It keeps a relaxed and biomechanically plausible posture with natural finger arches while preserving keystroke accuracy, and it reduces distortions.}
    \label{fig:apr_pipeline} 
\end{figure}
To bridge task accuracy and human-like execution without relying on expert datasets, we propose a bimanual framework with Adversarial Posture Regularization (APR). 

Our main contributions are:
\begin{itemize}
\item We propose \emph{Adversarial Posture Regularization} (APR), which effectively reduces unnatural postures in high-DoF dexterous hands during bimanual piano playing.
\item We develop a low-cost egocentric data collection method to capture piano hand demonstration data using a Meta Quest 3, and we release the resulting dataset publicly.
\item We introduce three metrics to quantify hand-motion naturalness, and demonstrate substantial improvements over the strongest baseline (PianoMime~\cite{qian2025pianomime}) on this benchmark across all three measures.
\end{itemize}

\section{RELATED WORK}
\subsection{Dexterous Manipulation and Piano-Playing Robots}

Controlling high-degree-of-freedom (DoF) dexterous hands remains one of the most formidable challenges in the field of robotics, primarily due to the curse of dimensionality and the complex, contact-rich dynamics involved in multi-finger coordination. Traditional approaches to dexterous manipulation have predominantly relied on trajectory optimization and model predictive control (MPC) to achieve stable grasping and precise object handling~\cite{chanrungmaneekul2023non,sleiman2019contact}. However, these analytical methods often struggle to scale to highly dynamic, continuous tasks.
In recent years, deep reinforcement learning (RL) has emerged as a powerful paradigm for tackling complex dexterous tasks, enabling high-dimensional policies to be learned directly through interaction in simulation. Meanwhile, techniques such as large-scale training and domain randomization can improve policy robustness and facilitate transfer to real-world settings. For example, OpenAI’s Dactyl system demonstrated dexterous in-hand object reorientation using the Shadow Hand~\cite{andrychowicz2020learning}.

Within the broader scope of dexterous control, robotic piano playing has recently been established as a premier benchmark. Unlike static grasping, piano playing requires high-frequency, multi-finger spatiotemporal coordination, stringent bimanual synchronization, and precise force application. RoboPianist~\cite{zakka2023robopianist} introduced a comprehensive RL framework and a high-fidelity MuJoCo benchmark to tackle this exact challenge, proving that dual-dexterous hands can learn to execute complex MIDI scores. Building upon this foundational environment~\cite{zakka2023robopianist}, PianoMime~\cite{qian2025pianomime} extended the RoboPianist repository by incorporating spatial tracking rewards derived from human demonstrations (extracted via monocular video). This framework serves as the direct software baseline and residual control architecture for our system~\cite{silver2018residual,johannink2018residual}. 

Beyond classic baselines, recent work has also advanced robotic piano playing through large-scale data and generative policy learning. RP1M~\cite{zhao2024rp1m} introduces a million-scale bimanual robot piano motion dataset with automatically annotated finger placements, enabling imitation learning to scale across diverse pieces. Building on this direction, OmniPianist~\cite{chen2025dexterous} performs large-scale training across thousands of piece-specific agents and aggregates their experience into an even larger trajectory corpus (RP1M++), yielding a single generalist policy capable of performing on the order of nearly one thousand pieces. Complementarily, PANDORA~\cite{huang2025pandora} explores diffusion-based policy learning for robotic piano performance, demonstrating that denoising sequence generation can produce high-dimensional action trajectories and achieve strong performance within the RoboPianist~\cite{zakka2023robopianist} environment. Although their playing accuracy continues to improve (e.g., higher keystroke-level F1-scores), these methods remain prone to ``reward hacking''~\cite{skalse2022defining}, where the learned policy exploits the simulator objectives and yields unnatural hand behaviors.
\subsection{Hand Motion Capture}

Collecting high-quality hand motion demonstrations remains a fundamental challenge for dexterous manipulation research. 
Previous work often relies on instrumented data gloves or marker-based optical motion capture systems to record human hand movements~\cite{gao2023hand}. 
Data gloves (e.g., CyberGlove) provide direct measurements of finger flexion through embedded sensors, enabling robust acquisition of high-frequency finger articulation with minimal sensitivity to occlusions~\cite{jarque2020large}. However, optical motion capture is costly and cumbersome to deploy, while wearing data gloves can hinder natural finger dexterity during piano performance.

Using inside-out tracking and egocentric cameras, VR systems allow scalable recording of natural hand movements without specialized motion-capture infrastructure. 
Recent work has also explored VR as an intuitive teleoperation interface, using commodity VR headsets (e.g., Oculus) to collect robot manipulation demonstrations through user-friendly interaction pipelines~\cite{george2025openvr}.
\subsection{Motion Priors and Style Regularization in RL}

Synthesizing natural, biologically plausible behaviors in RL agents has been a long-standing pursuit in character animation and robotics. Traditional imitation learning paradigms, such as Behavior Cloning (BC) and Generative Adversarial Imitation Learning (GAIL)~\cite{ho2016generative}, have been widely employed to transfer human skills to robots. While GAIL successfully introduced Generative Adversarial Networks (GANs) to match the state-action occupancy measure, it typically requires strictly aligned state-action pairs and heavily structured demonstration data, which are notoriously difficult to acquire for high-DoF dexterous systems.

To relax these strict data requirements and improve robustness, Adversarial Motion Priors (AMP)~\cite{peng2021amp} and its subsequent variants (e.g., ASE)~\cite{peng2022ase} pioneered the use of GAN-based distribution matching on observation-only state transitions. By decoupling the task objective from the style reward, these data-driven approaches have achieved tremendous success in synthesizing natural, stylized gaits for macroscopic full-body locomotion, such as bipedal humanoids and quadrupedal robots~\cite{peng2020learning}. 

However, extending humanoid-centric adversarial methods to dual dexterous hands (over 40 DoFs) remains largely unexplored. Unlike full-body locomotion, finger motion is constrained by intricate tendon-coupling biomechanics, and pure imitation learning for piano often disrupts the precise keystroke timing needed for musical accuracy.

\begin{figure*}[!t]
    \centering
    \includegraphics[width=0.95\textwidth]{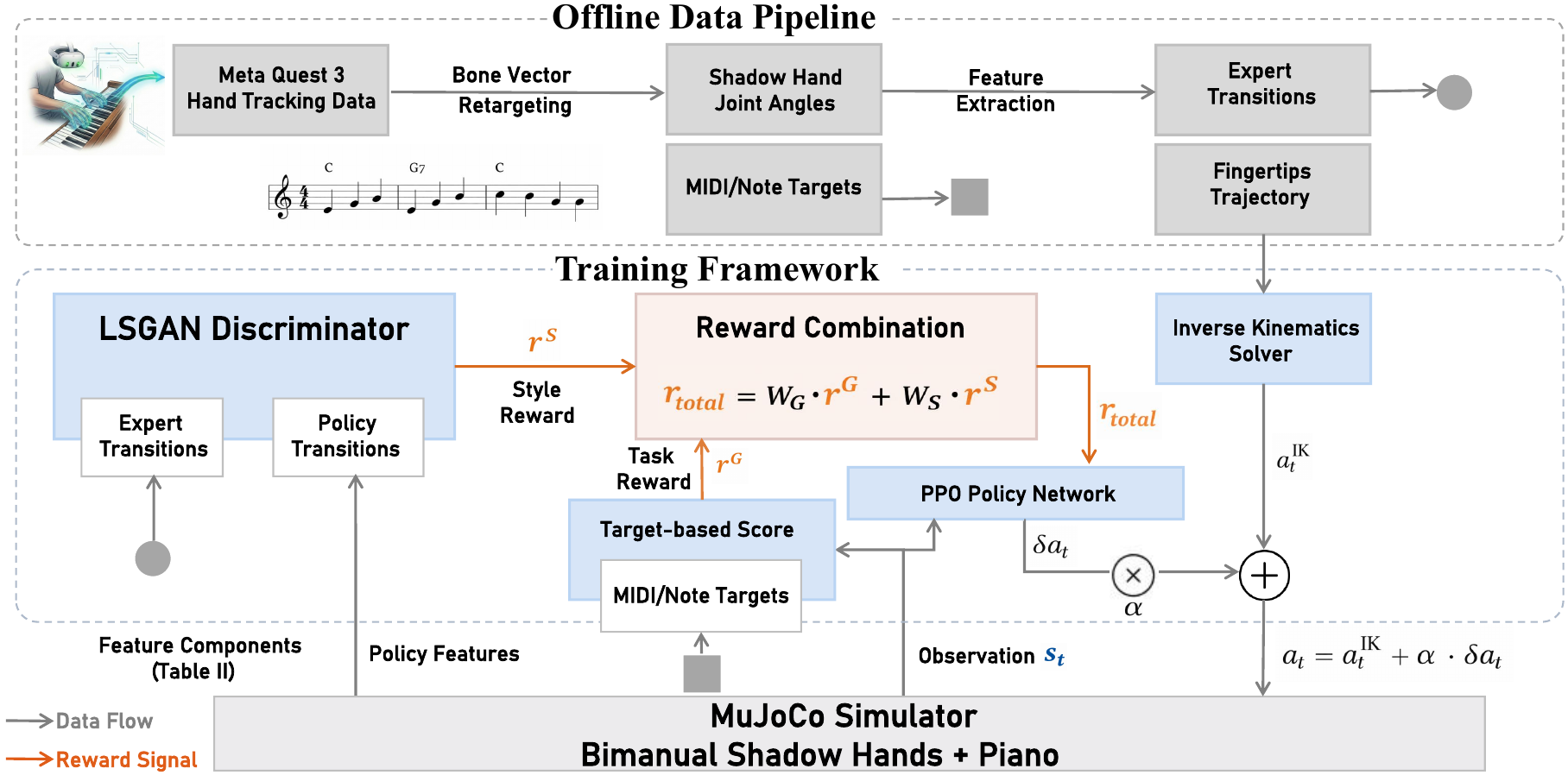}
    \caption{We first collect hand demonstrations and retarget them to Shadow Hand joint angles to build an expert transition dataset. During training, an LSGAN~\cite{mao2017least} discriminator compares policy transitions to expert transitions to provide a style reward, which is combined with the task reward. A PPO policy learns a residual action that is added on top of an IK controller, and the resulting bimanual piano-playing motions are executed in MuJoCo.}
    \label{fig:overview}
\end{figure*}

\section{METHODOLOGY}
\label{sec:methodology}

\subsection{Task Formulation and System Overview}
\label{sec:task_formulation}

Building upon the environments established by RoboPianist~\cite{zakka2023robopianist}, we formulate the dual-dexterous piano playing task as a discrete-time Markov Decision Process (MDP). The simulation operates in MuJoCo at a 20\,Hz control frequency.

\subsubsection{State and Action Space}
The agent controls two Shadow Dexterous Hands augmented with 3D prismatic forearms. The observation $\tilde{o}_t \in \mathcal{S}$ stacks four consecutive frames of the components listed in Table~I.
\begin{table}[htbp]
\centering
\caption{Observation components per control step.}
\label{tab:observation}
\begin{tabular}{lrl}
\toprule
\textbf{Component} & \textbf{Dim} & \textbf{Description} \\
\midrule
$q_{\mathrm{hand}}$ & $54$ & Joint positions of left and right hands ($27 \times 2$) \\
$s_{\mathrm{piano}}$ & $88$ & Depression state of all 88 piano keys \\
$s_{\mathrm{sustain}}$ & $1$ & Sustain pedal state \\
$g_{t:t+H}$ & $979$ & Lookahead goal: binary key targets, $H{=}10$ steps \\
$f_t$ & $10$ & Per-finger binary indicators (5 per hand) \\
$a^{\mathrm{IK}}_t$ & $46$ & Prior action from the IK controller \\
$d_{t:t+H}$ & $396$ & Demonstrator fingertip trajectories, $H{=}10$ steps \\
\bottomrule
\end{tabular}
\end{table}

To ease the exploration burden in high-dimensional continuous control, we adopt the residual learning framework proposed in~\cite{qian2025pianomime,silver2018residual,garcia2020physics}. At each step, an inverse-kinematics (IK) controller generates a nominal action $a^{\mathrm{IK}}_t$ based on pre-computed fingertip trajectories. The policy $\pi_\theta(a_t \mid \tilde{o}_t)$ then outputs a residual correction $\delta a_t \in \mathbb{R}^{47}$ (46 hand actuators + 1 sustain pedal), yielding the final applied action:
\begin{equation}
\label{eq:residual_action}
a_{t,1:46} = a^{\mathrm{IK}}_{t} + \alpha \cdot \delta a_{t,1:46}, \quad a_{t,47} = \delta a_{t,47}
\end{equation}
where $\alpha$ scales the residual, constraining the policy to fine-grained adjustments around the IK prior rather than learning gross hand positioning from scratch.

\subsubsection{Task Reward Function}
We directly inherit the task reward formulation $r^G(s_t, a_t)$ from RoboPianist, which comprises three main components:
\begin{equation}
\label{eq:task_reward}
r^G = r_{\mathrm{key}} + r_{\mathrm{sustain}} + r_{\mathrm{forearm}}
\end{equation}
Specifically, $r_{\mathrm{key}}$ applies a Gaussian-tolerance reward for accurately depressing the target keys and a binary penalty for striking incorrect keys (false positives). $r_{\mathrm{sustain}}$ evaluates the sustain pedal accuracy, and $r_{\mathrm{forearm}}$ penalizes collisions between the two forearms.

While the task reward $r^G$ effectively drives the F1-score for musical accuracy, it alone cannot rule out physically implausible behaviors; similarly, IK on redundant high-DoF hands can produce mathematically optimal but visually unnatural solutions. We therefore introduce Adversarial Posture Regularization (APR) as a style reward to enforce biomechanical naturalness during high-dynamic playing in~\ref{sec:apr}.

\subsection{VR-based Prior Acquisition and Retargeting}

A key contribution of our approach is that the human motion prior required by the APR discriminator is acquired using consumer-grade VR hardware rather than expensive optical motion capture systems. We describe the full pipeline from raw hand-tracking capture to the final reference dataset $\mathcal{D}$.

\subsubsection{Data Acquisition}

We collect egocentric hand-motion demonstrations using Meta Quest 3’s markerless hand tracking. The headset provides 3D joint positions $\mathbf{p}_j \in \mathbb{R}^3$ and orientations (quaternions) $\mathbf{q}_j \in \mathbb{H}$ for 26 joints per hand at ${\sim}30$ Hz, capturing both hands with timestamps.
\begin{figure}[t]
    \centering
    \includegraphics[width=\columnwidth]{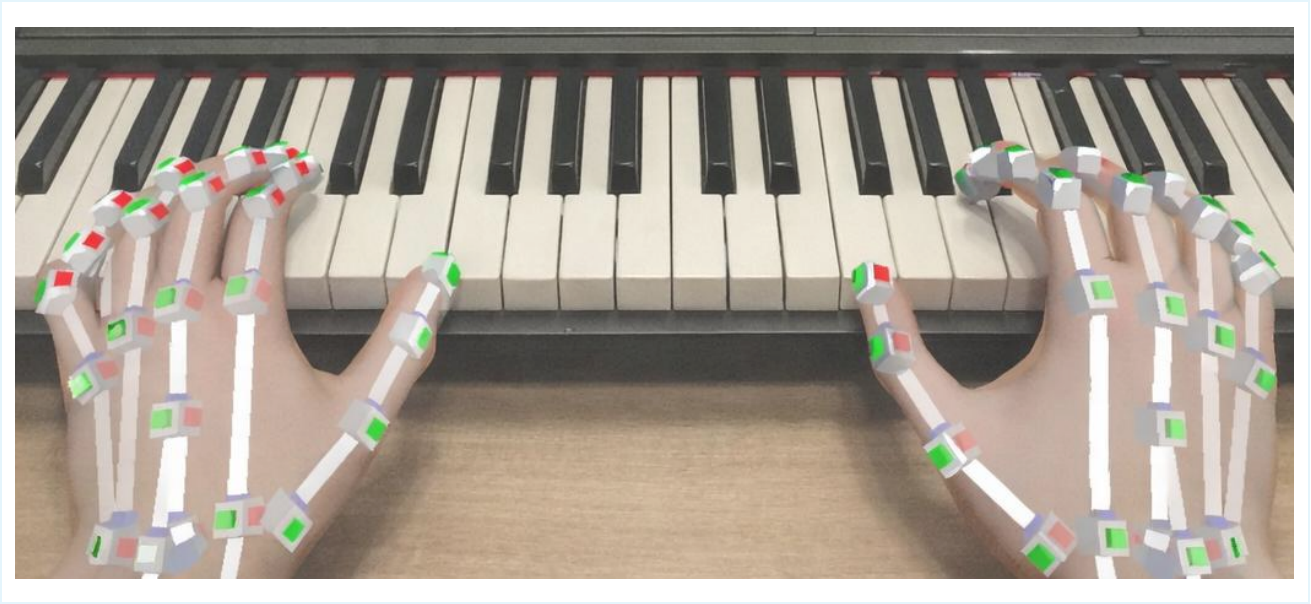} 
    \caption{Egocentric photo-capture setup for hand demonstration collection.}
    \label{fig:data_collection_setup}
\end{figure}
\label{sec:retargeting}

\subsubsection{Kinematic Retargeting}

Since the human hand and the Shadow Dexterous Hand differ in kinematic topology, segment lengths, and joint limits, we adopt a Vector Bone Retargeting approach that maps bone-direction angles rather than absolute positions, thus achieving morphology-invariant correspondence.

\paragraph{Palm frame construction}
For each frame, we construct a local palm coordinate system $\{\hat{\mathbf{f}}, \hat{\mathbf{l}}, \hat{\mathbf{n}}\}$:
\begin{equation}
\hat{\mathbf{f}} = \frac{\mathbf{p}_{\mathrm{Palm}} - \mathbf{p}_{\mathrm{Wrist}}}{\|\mathbf{p}_{\mathrm{Palm}} - \mathbf{p}_{\mathrm{Wrist}}\|}
\end{equation}
The lateral axis $\hat{\mathbf{l}}$ is derived from the knuckle line (IndexProximal $\to$ RingProximal), and the dorsal normal $\hat{\mathbf{n}}$ is obtained via cross products, with sign conventions adjusted for left/right handedness. This frame ensures invariance to global hand position and orientation.

\paragraph{Index/middle/ring/little finger retargeting}
For each of the four regular fingers, we define the bone chain from Metacarpal through Proximal, Intermediate, Distal, to Tip. The retargeted MCP, PIP, and DIP flexion angles are computed as the unsigned angle between consecutive bone vectors:
\begin{equation}
\label{eq:finger_retarget}
\begin{aligned}
\theta_{\mathrm{MCP}} &= \arccos\!\left(\frac{\vec{b}_0 \cdot \vec{b}_1}{\|\vec{b}_0\|\,\|\vec{b}_1\|}\right),\\
\theta_{\mathrm{PIP}} &= \arccos\!\left(\frac{\vec{b}_1 \cdot \vec{b}_2}{\|\vec{b}_1\|\,\|\vec{b}_2\|}\right),\\
\theta_{\mathrm{DIP}} &= \arccos\!\left(\frac{\vec{b}_2 \cdot \vec{b}_3}{\|\vec{b}_2\|\,\|\vec{b}_3\|}\right).
\end{aligned}
\end{equation}
where $\vec{b}_i = \mathbf{p}_{i+1} - \mathbf{p}_{i}$ are the bone direction vectors. Since the hand shape required by our task does not rely on lateral finger splay, we set the abduction/adduction angles to zero and exclude the corresponding joints from the APR feature space; this also helps prevent the discriminator from learning spurious correlations from low-resolution, noise-prone abduction signals in the tracker.

\paragraph{Wrist retargeting}
We compute the relative rotation from a neutral reference frame (first captured frame) to the current wrist orientation:
\begin{equation}
\Delta R = R_{\mathrm{ref}}^{-1} \circ R_{\mathrm{cur}}
\end{equation}
The relative rotation is decomposed into intrinsic YXZ Euler angles to extract ulnar/radial deviation (WRJ2, yaw component) and flexion/extension (WRJ1, pitch component), each attenuated by a factor of 0.8 to account for the Shadow Hand's more restricted wrist range of motion.

\paragraph{Thumb retargeting}
Retargeting the opposable thumb is challenging due to the mismatch between the human thumb base (saddle joint) and the Shadow Hand's 5-DoF thumb (THJ5--THJ1). To reduce sensitivity to global hand motion, we first express thumb bone directions in a palm-local coordinate frame:
\begin{equation}
\mathbf{b}^{L}_{i} =
\begin{bmatrix}
\hat{\mathbf{f}}^{\top}\mathbf{b}_{i} \\
\hat{\mathbf{l}}^{\top}\mathbf{b}_{i} \\
\hat{\mathbf{n}}^{\top}\mathbf{b}_{i}
\end{bmatrix}.
\end{equation}
We then map changes in in-plane orientation and elevation, together with proximal/distal inter-bone angles, to the Shadow Hand thumb joints (THJ5--THJ1) using a lightweight heuristic that yields visually plausible motions in simulation.

\subsubsection{Building the Reference Dataset}

The retargeted joint trajectories are converted into the APR discriminator's feature space. For each hand, we extract a \textbf{41-dimensional state descriptor} $\Phi_{\mathrm{hand}}(s)$:
\begin{equation}
\label{eq:feature_vector}
\Phi_{\mathrm{hand}}(s) = \left[\bar{\mathbf{q}}_{1:18},\; \bar{\dot{\mathbf{q}}}_{1:18},\; \bar{\mathbf{z}}_{1:5}\right] \in \mathbb{R}^{41},
\end{equation}
where $\bar{\mathbf{q}}$ are normalized joint angles, $\bar{\dot{\mathbf{q}}}$ are clipped angular velocities, and $\bar{\mathbf{z}}$ are clipped fingertip heights relative to the wrist (Table~\ref{tab:features}).

\begin{table}[htbp]
\centering
\caption{Per-hand feature vector components (41 dimensions).}
\label{tab:features}
\setlength{\tabcolsep}{6pt}
\renewcommand{\arraystretch}{1.15}
\small
\begin{tabular}{l c l}
\toprule
\textbf{Component} & \textbf{Dims} & \textbf{Description} \\
\midrule
$\bar{\mathbf{q}}_{1:18}$ & 18 & Joint angles (min--max norm.) \\
$\bar{\dot{\mathbf{q}}}_{1:18}$ & 18 & Angular velocities (clipped) \\
$\bar{\mathbf{z}}_{1:5}$ & 5 & Fingertip heights (clipped) \\
\bottomrule
\end{tabular}

\vspace{2pt}
\footnotesize
\textit{Normalisation:}
$\bar q_i=\frac{q_i-q_i^{\min}}{q_i^{\max}-q_i^{\min}}\in[0,1]$,\;
$\bar{\dot q}_i=\mathrm{clip}(\dot q_i/v^{\max},-1,1)$,\;
$\bar z_i=\mathrm{clip}((z_i^{\mathrm{tip}}-z^{\mathrm{wrist}})/0.1,-1,1)$.
\end{table}

The final reference dataset is constructed as consecutive state-transition pairs:
\begin{equation}
\label{eq:dataset}
\mathcal{D} = \left\{\left(\Phi_{\mathrm{hand}}(s_t),\; \Phi_{\mathrm{hand}}(s_{t+1})\right)\right\}_{t=1}^{T-1}.
\end{equation}
We pool samples from both hands into a single buffer and use a shared-weight discriminator
$[\Phi_{\mathrm{hand}}(s_t),\, \Phi_{\mathrm{hand}}(s_{t+1})] \in \mathbb{R}^{82}$.
\subsection{Adversarial Posture Regularization (APR)}
\label{sec:apr}

\subsubsection{Core Idea}

Inspired by the GAN paradigm~\cite{goodfellow2014generative} and Adversarial Motion Priors~\cite{peng2021amp}, we propose \textbf{Adversarial Posture Regularization (APR)}. APR reformulates the naturalness objective as \emph{distribution matching}: using a small amount of reference data (Sec.~\ref{sec:retargeting}) to replace per-frame tracking that requires high-precision 3D hand pose. Crucially, the discriminator does not require strict temporal alignment between the policy and the reference. Instead, it simply asks, ``does this state transition look like it comes from a human pianist?'' This distributional view is robust to the frame-wise noise inherent in Meta Quest~3 hand tracking: while individual frames can be jittery, the overall statistical structure of human piano motion (e.g., coupled joint flexion patterns, smooth velocity profiles, and natural fingertip-height envelopes) is largely preserved. Concretely, APR online trains a discriminator network $D_\phi$ to distinguish between the \emph{distribution} of state transitions generated by the RL policy $\pi_\theta$ and that observed in a human reference dataset $\mathcal{D}$. 
As a result, APR reduces unnatural postures, remains robust to noisy measurements, and can be reused across different songs without modification.

\subsubsection{Shared-Weight Bimanual Architecture}

A key architectural design of APR is the use of a \textbf{single discriminator with shared weights} that processes left-hand and right-hand state transitions identically. Let $\Phi_L(s_t) \in \mathbb{R}^{41}$ and $\Phi_R(s_t) \in \mathbb{R}^{41}$ denote the per-hand feature descriptors (Eq.~\ref{eq:feature_vector}). Rather than training two separate discriminators, we define a single network $D_\phi: \mathbb{R}^{82} \to \mathbb{R}$ that takes the concatenated transition as input and apply it independently to each hand:
\begin{equation}
\label{eq:shared_disc}
\begin{aligned}
d_L &= D_\phi\!\left([\Phi_L(s_t),\, \Phi_L(s_{t+1})]\right), \\
d_R &= D_\phi\!\left([\Phi_R(s_t),\, \Phi_R(s_{t+1})]\right).
\end{aligned}
\end{equation}
This design encourages bilateral symmetry and improves sample efficiency.

\subsubsection{Discriminator Network}

The discriminator $D_\phi$ is a fully-connected MLP:
\begin{equation}
D_\phi: \mathbb{R}^{82} \xrightarrow{\mathrm{FC}(1024)} \mathrm{ReLU} \xrightarrow{\mathrm{FC}(256)} \mathrm{ReLU} \xrightarrow{\mathrm{FC}(1)} \mathbb{R}
\end{equation}
The output is an unbounded scalar logit (no sigmoid or tanh activation), following the Least-Squares GAN (LSGAN) convention. All weights are initialised with orthogonal initialisation (gain $\sqrt{2}$ for hidden layers, gain 1.0 for the output head).

\subsubsection{Loss Function}

We train the discriminator with the Least-Squares GAN (LSGAN) objective~\cite{mao2017least}. The discriminator loss is:
\begin{equation}
\label{eq:lsgan}
\begin{aligned}
\mathcal{L}_D
&= \underbrace{\mathbb{E}_{(s,s') \sim \mathcal{D}}
\!\left[\left(D_\phi(s,s') - 1\right)^2\right]}_{\text{expert}\to +1} \\
&\quad + \underbrace{\mathbb{E}_{(s,s') \sim \pi_\theta}
\!\left[\left(D_\phi(s,s') + 1\right)^2\right]}_{\text{policy}\to -1}
+ \mathcal{L}_{\mathrm{GP}} .
\end{aligned}
\end{equation}
where we use the notational shorthand $D_\phi(s,s') \triangleq D_\phi([\Phi_{\mathrm{hand}}(s),\, \Phi_{\mathrm{hand}}(s')])$. The discriminator is trained to output $+1$ for expert transitions and $-1$ for policy-generated transitions.

To prevent the discriminator from becoming overly confident and collapsing the reward signal, we apply a \textbf{gradient penalty} regulariser on the expert data:
\begin{equation}
\label{eq:grad_penalty}
\mathcal{L}_{\mathrm{GP}} = \frac{w_{\mathrm{gp}}}{2}\, \mathbb{E}_{(s,s') \sim \mathcal{D}} \left[\left\|\nabla_\Phi D_\phi(s,s')\right\|^2\right]
\end{equation}
 This penalty encourages the discriminator to be Lipschitz-smooth in the neighbourhood of expert data, stabilising the adversarial training dynamics and producing a well-shaped reward landscape.

The discriminator is updated online once per policy rollout using minibatches sampled from an expert buffer and a replay buffer of policy transitions.

\subsection{Hybrid Reward and Optimisation}
\label{sec:hybrid_reward}

The agent must balance task performance (pressing the correct keys at the correct times) and style quality (natural hand posture and motion). We achieve this with a hybrid reward and a structured optimisation loop.

\subsubsection{Style Reward}

The style reward is the core output of the APR mechanism. It is derived directly from the discriminator's assessment of how ``human-like'' the current state transition appears. For each hand $h \in \{L, R\}$, the per-hand style reward is:
\begin{equation}
\label{eq:style_reward}
r^S_h(s_t, s_{t+1}) = \max\!\left(0,\; 1 - \lambda \cdot \left(D_\phi(s_t, s_{t+1})_h - 1\right)^2\right)
\end{equation}
where $\lambda$ is a scaling hyperparameter that controls the reward sharpness.
The combined style reward averages over both hands:
\begin{equation}
\label{eq:style_combined}
r^S_t = \frac{1}{2}\left(r^S_L(s_t, s_{t+1}) + r^S_R(s_t, s_{t+1})\right)
\end{equation}
This averaging prevents the policy from sacrificing naturalness in one hand to improve the other.

\subsubsection{Joint Optimisation}

The total reward presented to the PPO optimiser is the weighted sum:
\begin{equation}
\label{eq:total_reward}
r_{\mathrm{total},t} = w_G \cdot r^G_t + w_S \cdot r^S_t
\end{equation}
where $w_G$ and $w_S$ are task and style weights respectively. The weights control the trade-off between task accuracy and motion naturalness: larger $w_S$ emphasises natural-looking motion, while larger $w_G$ emphasises key-press accuracy.

\paragraph{Policy optimisation}
We optimise an actor--critic policy with \textbf{PPO}~\cite{schulman2017proximal} using the clipped surrogate objective:
\begin{equation}
\label{eq:ppo}
\begin{aligned}
\mathcal{L}^{\mathrm{PPO}}_\theta
= -\mathbb{E}_t\Big[
&\min\Big(
\frac{\pi_\theta(a_t|s_t)}{\pi_{\theta_{\mathrm{old}}}(a_t|s_t)}\hat{A}_t, \\
&\mathrm{clip}\!\Big(
\frac{\pi_\theta(a_t|s_t)}{\pi_{\theta_{\mathrm{old}}}(a_t|s_t)},
\, 1 \pm \epsilon_{\mathrm{clip}}
\Big)\hat{A}_t
\Big)
\Big]
\end{aligned}
\end{equation}
with $\epsilon_{\mathrm{clip}} = 0.2$.

\paragraph{Advantage recomputation}
Since the style reward is computed after discriminator evaluation,
we recompute the GAE advantages using the hybrid reward before PPO updates.

\paragraph{Alternating optimisation loop}
The full training procedure alternates between three phases per iteration:
\begin{enumerate}

    \item \textbf{Rollout collection.} The policy $\pi_\theta$ collects on-policy transitions by interacting with $N_{\mathrm{env}}$ parallel MuJoCo environments. Per-hand APR features $\Phi_L, \Phi_R$ are extracted at each step by the environment wrapper.
    \item \textbf{APR update.} Transitions are added to a shared replay buffer, and the discriminator $D_\phi$ is updated using the LSGAN objective with gradient penalty (Eq.~\ref{eq:lsgan}). The resulting style rewards are computed for the rollout and combined with the task reward (Eq.~\ref{eq:total_reward}).
    \item \textbf{PPO update.} The policy $\pi_\theta$ and value function $V_\psi$ are then optimised with PPO using the hybrid reward.
\end{enumerate}

\section{EXPERIMENTS}
\label{sec:experiments}

We evaluate the proposed APR framework on a suite of piano pieces of varying difficulty, comparing against two baselines that represent the dominant paradigms in prior work. Our experiments address three questions: (1)~Does APR preserve task performance (note accuracy) compared to baselines? (2)~Does APR produce measurably more natural hand motion? (3)~What is the data efficiency of the distributional matching approach relative to trajectory-tracking alternatives?

\subsection{Experimental Setup}
\label{sec:exp_setup}

\subsubsection{Simulation Environment}

All experiments are conducted in MuJoCo using a pair of Shadow Dexterous Hands (24 joints, 20 tendon-coupled actuators, 3 forearm DOFs each) positioned above a full 88-key digital piano. The control frequency is 20\,Hz ($\Delta t = 0.05$\,s) with a physics timestep of 0.005\,s. Each policy receives a 4-frame stacked observation and outputs a 47-dimensional residual action (Sec.~\ref{sec:task_formulation}).

\subsubsection{APR Reference Data}

A key advantage of our approach is its profound data efficiency. The entire APR reference dataset $\mathcal{D}$ is captured in a single, short session using a consumer-grade Meta Quest~3 headset, requiring no optical markers, no multicamera calibration, and no specialized lab infrastructure. The raw dataset is detailed in Table~\ref{tab:expert_data}.

\begin{table}[htbp]
\centering
\caption{APR reference data collection.}
\label{tab:expert_data}
\begin{tabular}{lrrr}
\toprule
\textbf{Clip} & \textbf{Frames} & \textbf{Duration (s)} & \textbf{Transitions} \\
\midrule
Free improvisation  & 374 & 18.7 & 373 \\
\bottomrule
\end{tabular}
\end{table}

Crucially, this incredibly brief reference clip is \emph{not aligned to any specific musical piece}; the exact same 18.7\,s dataset is universally reused to regularize the policy across all target test songs without modification. This stands in stark contrast to explicit tracking or DeepMimic-style approaches~\cite{peng2018deepmimic}, which require meticulously engineered, per-song, frame-by-frame aligned state-action demonstrations, making data preparation prohibitively expensive and fundamentally unscalable for new musical repertoires.

\subsubsection{Baselines}

We compare against two baselines that represent the dominant paradigms in prior work on dexterous piano playing:

\paragraph{2D Fingertip Tracking (PianoMime)}
The approach of PianoMime~\cite{qian2025pianomime}, which extracts 2D fingertip positions from video and uses explicit per-frame tracking objectives to drive the hand toward reference fingertip locations. This baseline has access to higher-quality reference data (video-derived 2D keypoints) but can only constrain fingertip positions.

\paragraph{Ours (APR)}
Our full method: residual PPO with Adversarial Posture Regularization using Meta Quest~3 reference data.

All methods share the same network architecture (FC 1024--256), training hyperparameters (24 parallel environments, 512 steps per rollout, 2000 training iterations, learning rate $3 \times 10^{-4}$ with exponential decay 0.999), and residual IK controller ($\alpha = 0.03$).

\subsubsection{Evaluation Metrics}

We evaluate along two orthogonal axes: \emph{task performance} and \emph{motion naturalness}.

\paragraph{Task performance}
We report the \textbf{F1 score}, the harmonic mean of precision and recall over per-timestep key activations:
\begin{equation}
\mathrm{F1} = \frac{2 \cdot \mathrm{Precision} \cdot \mathrm{Recall}}{\mathrm{Precision} + \mathrm{Recall}}
\end{equation}
where Precision measures the fraction of activated keys that should be pressed, and Recall measures the fraction of target keys that are actually pressed.

\paragraph{Motion naturalness}
We propose three complementary biomechanical metrics that quantify different aspects of hand naturalness, evaluated over the full trajectory of raw joint angles $q \in \mathbb{R}^{18}$:

\begin{itemize}
    \item \textbf{cPSI} (Continuous Posture Strain Index): Measures static posture deviation from the biomechanical neutral position. For each joint $i$ with range $[q_i^{\min}, q_i^{\max}]$:
    \begin{equation}
    \mathrm{cPSI} = \frac{1}{T \cdot D} \sum_{t=1}^{T} \sum_{i=1}^{D} \left(\frac{2q_{t,i} - (q_i^{\max} + q_i^{\min})}{q_i^{\max} - q_i^{\min}}\right)^2
    \end{equation}
    This metric quantifies excessive joint strain by measuring how far joints deviate from their neutral posture, particularly capturing extreme hyper-extension.
    \item \textbf{BSE} (Biomechanical Synergy Error): Measures violation of the natural tendon coupling between DIP and PIP joints. In the human hand, the flexor digitorum profundus tendon enforces an approximate coupling $q_{\mathrm{DIP}} \approx \frac{2}{3} q_{\mathrm{PIP}}$~\cite{infantino2003human}. We measure deviation from this synergy across the four fingers:
    \begin{equation}
    \mathrm{BSE} = \frac{1}{T \cdot 4} \sum_{t=1}^{T} \sum_{f=1}^{4} \left(q_{\mathrm{DIP}}^{(f)} - \tfrac{2}{3}\, q_{\mathrm{PIP}}^{(f)}\right)^2
    \end{equation}
    This metric specifically detects the \emph{``zombie hand''} artefact of 2D fingertip tracking.

    \item \textbf{FAC} (Finger Arc Continuity): Measures the smoothness of the finger profile from MCP through PIP to DIP using the discrete Laplacian (second-order finite difference), plus a penalty for hyperextension:
    \begin{equation}
    \begin{aligned}
    \mathrm{FAC} = 
    \frac{1}{T \cdot 4} 
    &\sum_{t=1}^{T} \sum_{f=1}^{4} \Bigg[
    \left(q_{\mathrm{MCP}}^{(f)} - 2q_{\mathrm{PIP}}^{(f)} + q_{\mathrm{DIP}}^{(f)}\right)^2 \\
    &+ \lambda \sum_{j \in \{M,P,D\}} \max(-q_j, 0)^2
    \Bigg]
    \end{aligned}
    \end{equation}
    with $\lambda = 0.1$. This captures the overall ``gracefulness'' of the hand.
\end{itemize}

For all three naturalness metrics, \textbf{lower is better} (closer to natural human motion). We report the average of left-hand and right-hand scores.

\begin{figure}[!htbp]
    \centering
    \includegraphics[width=1\columnwidth]{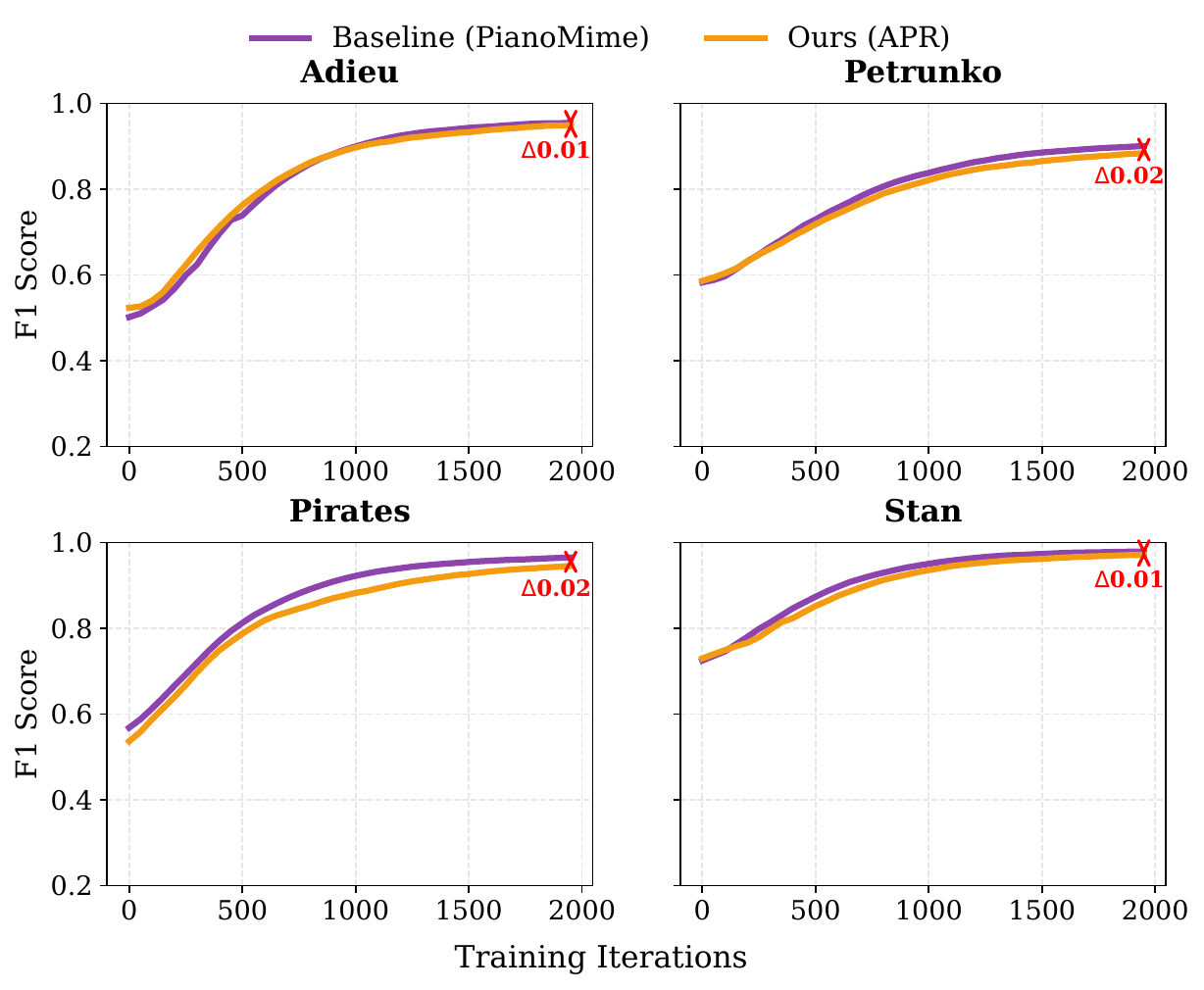}
    \caption{Task performance (F1 score) learning curves over training iterations for four musical tracks (Adieu, Petrunko, Pirates, Stan). APR (orange) converges similarly to the PianoMime baseline (purple) and achieves comparable final performance. The final performance gap $\Delta$, annotated in each plot, is small ($\approx 0.01$--$0.02$).}
    \label{fig:f1_curves}
\end{figure}
\subsection{Results}
\label{sec:results}

\subsubsection{Task Performance (F1 Score)}
Fig.~\ref{fig:f1_curves} plots the F1 learning curves over training iterations for all methods. Throughout training, APR’s F1 consistently tracks and closely approaches the 2D fingertip tracking baseline (PianoMime), and it achieves a comparable final F1 while also optimizing motion naturalness. This suggests that APR’s distributional matching objective acts as a soft regularizer rather than a per frame constraint, allowing the policy to learn accurate key pressing strategies while remaining within the manifold of natural hand motions.

\subsubsection{Qualitative Results and Quantitative Evaluation}
\begin{figure}[!htbp]
    \centering
    \includegraphics[width=0.98\columnwidth]{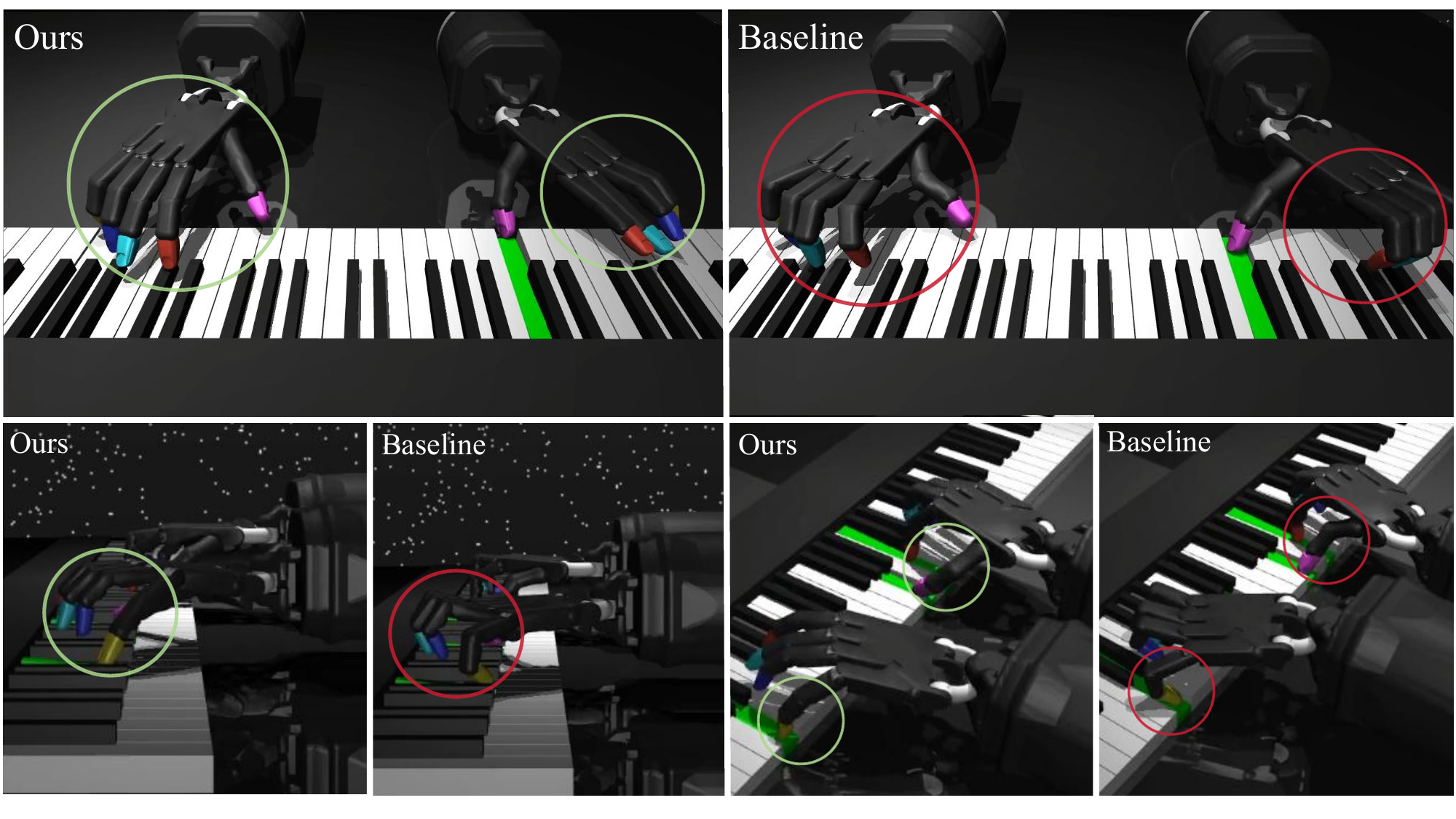}
    \caption{Green circles highlight natural, human-like hand motions, while red circles indicate unnatural postures or artifacts.}
    \label{fig:gesture}
\end{figure}

Table~\ref{tab:main_results} summarises performance across test songs. APR matches PianoMime on task accuracy (F1), suggesting that posture regularization does not meaningfully affect keystroke timing or bimanual coordination. Notably, the baseline relies heavily on an IK prior, which on the high-DoF and redundant Shadow Hand can yield solutions that are mathematically optimal yet visually inconsistent with natural human piano-playing gestures. In contrast, APR incorporates human demonstration priors and imposes a softer constraint over the IK solution space. As a result, APR consistently improves motion naturalness, with lower cPSI/BSE/FAC scores indicating reduced posture strain and fewer kinematic artifacts. Fig.~\ref{fig:gesture} provides qualitative evidence: APR produces more natural, human-like motions (green), whereas the baseline exhibits more unnatural postures or artifacts (red).
\subsubsection{Analysis}
This indicates that the distributional matching objective functions as a soft regularizer, shaping the policy toward human-like kinematics without over-constraining the control space. As a result, the agent can still discover accurate key-pressing strategies while avoiding posture strain and visually implausible IK solutions.
\begin{table}[!htbp]
\centering
\caption{Quantitative comparison across test songs.}
\label{tab:main_results}
\begin{tabular}{llcccc}
\toprule
\textbf{Song} & \textbf{Method} & \textbf{F1} & \textbf{cPSI} & \textbf{BSE} & \textbf{FAC} \\
\midrule

\multirow{3}{*}{\texttt{Adieu}}
    & PianoMime            & 0.962  & 0.3162  & 0.1627  & 1.7838  \\
    & \textbf{Ours (APR)}  & 0.960  & 0.2306  & 0.0929  & 0.9266  \\
    & $\Delta$ (\%)        & -0.2\% & -27.1\% & -42.9\% & -48.1\% \\
\midrule   
\multirow{3}{*}{\texttt{Petrunko}}
    & PianoMime            & 0.909  & 0.3460  & 0.4029  & 1.4157  \\
    & \textbf{Ours (APR)}  & 0.894  & 0.2674  & 0.1899  & 1.1702  \\
    & $\Delta$ (\%)        & -1.7\% & -22.7\% & -52.9\% & -17.3\% \\
\midrule
\multirow{3}{*}{\texttt{Pirates}}
    & PianoMime            & 0.969  & 0.3633  & 0.1369  & 2.4645  \\
    & \textbf{Ours (APR)}  & 0.954  & 0.2302  & 0.0829  & 1.2692  \\
    & $\Delta$ (\%)        & -1.5\% & -36.6\% & -39.4\% & -48.5\% \\
\midrule
\multirow{3}{*}{\texttt{Stan}}
    & PianoMime            & 0.982  & 0.3513  & 0.3924  & 1.4066  \\
    & \textbf{Ours (APR)}  & 0.968  & 0.2906  & 0.1457  & 1.3527  \\
    & $\Delta$ (\%)        & -1.4\% & -17.3\% & -62.9\% & -3.8\%  \\

\midrule
\bottomrule
\end{tabular}
\end{table}

\begin{table}[!t]
\centering
\caption{Data requirements comparison. }
\label{tab:data_comparison}
\begin{tabular}{lccc}
\toprule
 & \textbf{DeepMimic} & \textbf{APR (Ours)} \\
\midrule
Per-song alignment & Required & \textbf{Not needed} \\
Reference duration & Full song & \textbf{$\sim$18.7\,s total} \\
Transition pairs & --- & \textbf{746 total} \\
Reusable across songs & No & \textbf{Yes} \\
\bottomrule
\end{tabular}
\end{table}
\subsection{Data Efficiency}
\label{sec:data_efficiency}

The baseline (PianoMime) originally tried to use DeepMimic to improve hand posture. However, DeepMimic requires carefully time-aligned state action pairs for each target song, and collecting song-aligned 3D tracking data of piano playing hands is very expensive. As a result, they fell back to tracking 2D fingertip positions from videos, which does not work well. In contrast, APR uses distributional matching and only needs reference data sampled from the same distribution as natural piano playing. It does not require temporal alignment, per-song correspondence, or precise retargeting. Our entire reference dataset is about 19 seconds of casual playing, and it is reused unchanged for all test songs without any modification or per-song tuning.

A comparison between the two approaches is summarised in Table~\ref{tab:data_comparison}.

\section{CONCLUSION}
We propose Adversarial Posture Regularization (APR) for high-DoF bimanual piano playing to address kinematic redundancy in dexterous manipulation, where standard IK solvers and task-only RL often exploit the Shadow Hand’s large action space and produce unnatural, biomechanically implausible postures such as severe joint hyperextension. We also collect and release an egocentric, first-person VR dataset of piano-hand motions for APR, and introduce three physics-based metrics to directly quantify kinematic naturalness for more rigorous evaluation of stylistic imitation.

Our approach has two main limitations: the LSGAN-based discriminator can be sensitive to hyperparameters, requiring careful balancing of gradient penalties and the style reward weight, and the hard IK task prior can dominate during extremely fast and difficult passages, where timing demands may temporarily override the soft adversarial constraint and cause brief drops in posture quality. Future work will adapt the style weight to local musical complexity and explore more stable generative priors,  to improve training stability and high-speed robustness.

\bibliographystyle{IEEEtran}
\bibliography{ref}
\end{document}